\documentclass[10pt,twocolumn,letterpaper]{article}

\usepackage[pagenumbers]{cvpr} 

\usepackage[dvipsnames]{xcolor}

\definecolor{cvprblue}{rgb}{0.21,0.49,0.74}
\usepackage[pagebackref,breaklinks,colorlinks,citecolor=cvprblue]{hyperref}
\usepackage{amsmath}
\PassOptionsToPackage{hyphens}{url}
\usepackage{hyperref}

\def\logo{\makebox[0pt][l]{\hspace{0pt}\raisebox{-0.5ex}{\includegraphics[scale=0.02]{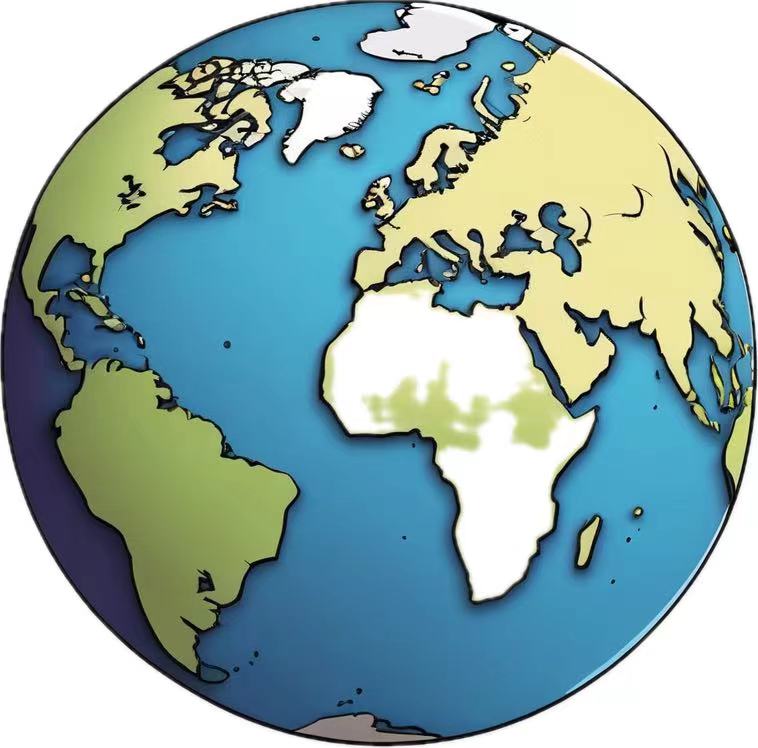}}}}

\def\pytorch{\makebox[0pt][c]{\hspace{0pt}\raisebox{-0.5ex}{\includegraphics[width=25pt]{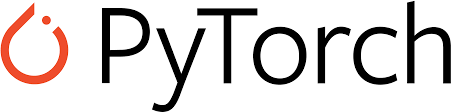}}}}

\def\paddle{\makebox[0pt][c]{\hspace{0pt}\raisebox{-0.5ex}{\includegraphics[width=20pt]{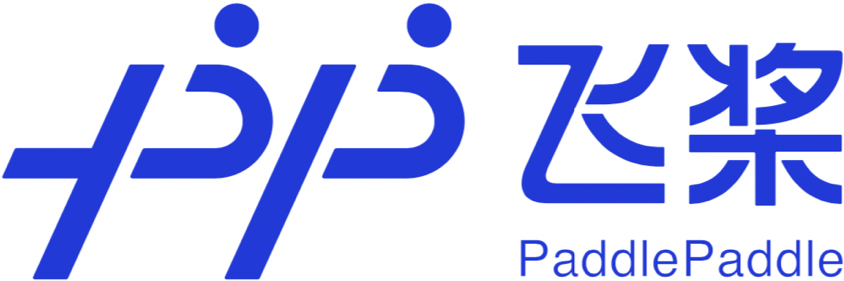}}}}

\def\matlab{\makebox[0pt][c]{\hspace{0pt}\raisebox{-0.5ex}{\includegraphics[width=25pt]{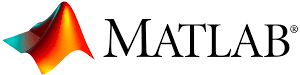}}}}

\def\paperID{*****} 
\def\confName{CVPR}
\def\confYear{2024}

\title{\ \ \!$_{\!}$\!\logo \ \ \ \ \! pen-CD: A Comprehensive Toolbox for Change Detection}

\author{Kaiyu Li\thanks{Equal contribution.}\\
Xi'an Jiaotong University\\
{\tt\small likyoo.ai@gmail.com}
\and
Jiawei Jiang$^{*}$\\
Sun Yat-Sen University\\
{\tt\small jiangjw26@mail2.sysu.edu.cn}
\and
Andrea Codegoni$^{*}$\\
University of Pavia\\
{\tt\small andrea.codegoni01@ateneopv.it}
\and
Chengxi Han$^{*}$\\
Wuhan University\\
{\tt\small chengxihan@whu.edu.cn}
\and
Yupeng Deng$^{*}$\\
Chinese Academy of Sciences\\
{\tt\small dengyp@aircas.ac.cn}
\and
Keyan Chen$^{*}$\\
Beihang University\\
{\tt\small kychen@buaa.edu.cn}
\and
Zhuo Zheng$^{*}$\\
Stanford University\\
{\tt\small zhuozheng@cs.stanford.edu}
\and
Hao Chen$^{*}$\\
Shanghai AI Laboratory\\
{\tt\small chenhao1@pjlab.org.cn}
\and
Ziyuan Liu$^{*}$\\
Tsinghua University \\
{\tt\small liuziyua22@mails.tsinghua.edu.cn}
\and
Yuantao Gu\\
Tsinghua University \\
{\tt\small gyt@tsinghua.edu.cn}
\and
Zhengxia Zou\\
Beihang University\\
{\tt\small zhengxiazou@buaa.edu.cn}
\and
Zhenwei Shi\\
Beihang University\\
{\tt\small shizhenwei@buaa.edu.cn}
\and
Sheng Fang\\
Shandong University of Science and Technology\\
{\tt\small fangsheng@tsinghua.org.cn}
\and
Deyu Meng\\
Xi'an Jiaotong University\\
{\tt\small dymeng@mail.xjtu.edu.cn}
\and
Zhi Wang\\
Xi'an Jiaotong University\\
{\tt\small zhiwang@xjtu.edu.cn}
\and
Xiangyong Cao\\
Xi'an Jiaotong University\\
{\tt\small caoxiangyong@mail.xjtu.edu.cn}
}

\begin{document}
\maketitle
\begin{abstract}
We present Open-CD, a change detection toolbox that contains a rich set of change detection methods as well as related components and modules. The toolbox started from a series of open source general vision task tools, including OpenMMLab Toolkits\footnote{\url{https://openmmlab.com}}, PyTorch Image Models\footnote{\url{https://github.com/huggingface/pytorch-image-models}}, etc. It gradually evolves into a unified platform that covers many popular change detection methods and contemporary modules. It not only includes training and inference codes, but also provides some useful scripts for data analysis. We believe this toolbox is by far the most complete change detection toolbox. In this report, we introduce the various features, supported methods and applications of Open-CD. In addition, we also conduct a benchmarking study on different methods and components. We wish that the toolbox and benchmark could serve the growing research community by providing a flexible toolkit to reimplement existing methods and develop their own new change detectors. Code and models are available at \url{https://github.com/likyoo/open-cd}. Pioneeringly, this report also includes brief descriptions of the algorithms supported in Open-CD, mainly contributed by their authors\footnote{Open-CD TRP: \url{https://github.com/likyoo/open-cd/tree/main/projects/open-cd\_technical\_report}}. We sincerely encourage researchers in this field to participate in this project and work together to create a more open community. \textbf{{\color{green!50!black}This toolkit and report will be kept updated.}}
\end{abstract}    
\section{Introduction}
\label{sec:intro}
Change detection is a fundamental remote sensing image interpretation task. The pipeline of change detection frameworks is usually similar to image segmentation tasks. But in particular, there is a pair of images as input, and the task focuses on detecting pixel-level differences in the bi-temporal images. Therefore, change detection is more complicated than single-temporal segmentation. And like other high-level vision tasks, different implementation settings can lead to very different results. Towards the goal of providing a high
quality codebase and unified benchmark, we build Open-CD, an open-source change detection codebase.

Major features of Open-CD are: (1) \textbf{Reliable dependencies.} Open-CD is built on the OpenMMLab Toolkits, especially MMCV\footnote{\url{https://github.com/open-mmlab/mmcv}}, MMEngine\footnote{\url{https://github.com/open-mmlab/mmengine}}, MMPretrain\footnote{\url{https://github.com/open-mmlab/mmpretrain}}, MMSegmentation\footnote{\url{https://github.com/open-mmlab/mmsegmentation}}, and MMDetection\footnote{\url{https://github.com/open-mmlab/mmdetection}} \cite{chen2019mmdetection}, and can arbitrarily call their components, including pipeline, model, module, loss, data augmentation, \etc, in the config file. (2) \textbf{Modular design}. We decompose the change detection framework into different components and users can easily construct a customized change detection method by combining different modules. (3) \textbf{Support of multiple methods out of box.} Open-CD supports typical and popular change detection methods, see Section \ref{sec:supported_methods} for the full list. (4) \textbf{State of the art.} In most cases, the Open-CD implementation performs better than the official one. It also helps some users win change detection challenges. (5) \textbf{High efficiency.} Depending on efficient MMEngine and MMCV, the training speed of Open-CD is faster than or comparable to other codebases.

In this report, we introduce the composition and workflow of the Open-CD toolbox, and conduct a benchmarking study on the typical or state-of-the-art change detection algorithms, including parameter, inference speed, etc. Apart from the above, this report provides a brief description for each of the supported algorithms in Open-CD. Specifically, we have launched the Open-CD Technical Report Plan (\textbf{{\color{green!50!black}Open-CD TRP}} for shot). We invite some authors  to introduce their algorithms and participate in the construction of the Open-CD codebase. \textbf{{\color{green!50!black}This plan is under active development and we will keep this report updated.}}
 
The rest of this report is organized as follows. We first introduce various supported methods, followed by a focus on features of Open-CD. Then, we present the benchmark results. Finally, we give examples of application scenarios of Open-CD, especially in the era of foundation models.

\begin{table*}
  \caption{Supported features of different change deetction codebases. ``\checkmark'' means officially supported and blank means not supported.}
  \label{table_1}
  \centering
  \scalebox{0.85}{
  \begin{tabular}{@{}lccccc@{}}
    \toprule[1pt]
    & Open-CD & change\_detection.pytorch\footnotemark & ChangeDetectionRepository\footnotemark & CDLab\footnotemark & ChangeDetectionToolbox\footnotemark \\
    & \pytorch & \pytorch & \pytorch & \paddle & \matlab \\
    \midrule[1pt]
    \textbf{\textit{Tasks:}} \\
    Binary & \checkmark & \checkmark & \checkmark & \checkmark & \checkmark \\
    Semantic & \checkmark & & & & \\
    \midrule[1pt]
    \textbf{\textit{Customized Models:}} \\
    Classification Nets & \checkmark & \checkmark & & & \\
    Segmentation Nets & \checkmark & \checkmark & & & \\
    \midrule[1pt]
    \textbf{\textit{Traditional Methods:}} \\
    CVA~\cite{malila1980change} & & & \checkmark & & \checkmark \\
    DPCA~\cite{deng2008pca} & & & & & \checkmark \\
    ImageDiff~\cite{mahmoudzadeh2007digital} & & & & & \checkmark \\
    ImageRatio~\cite{lu2004change} & & & & & \checkmark \\
    ImageRegr~\cite{lu2004change} & & & & & \checkmark \\
    MAD~\cite{nielsen1998multivariate} & & & \checkmark & & \checkmark \\
    IRMAD~\cite{nielsen2007regularized} & & & & & \checkmark \\
    PCA-Kmeans~\cite{celik2009unsupervised} & & & \checkmark & & \checkmark \\
    PCDA~\cite{zhang2007remote} & & & & & \checkmark \\
    \midrule[1pt]
    \textbf{\textit{Deep Learning Methods:}} \\
    FC-EF~\cite{daudt2018fully} & \checkmark & & & \checkmark & \\
    FC-Siam-Diff~\cite{daudt2018fully} & \checkmark & & & \checkmark & \\
    FC-Siam-Conc~\cite{daudt2018fully} & \checkmark & & & \checkmark & \\
    STANet~\cite{chen2020spatial} & \checkmark & \checkmark & & \checkmark & \\
    DSIFN~\cite{zhang2020deeply} & \checkmark & & & \checkmark \\
    L-UNet~\cite{papadomanolaki2021deep} & & & & \checkmark & \\
    SNUNet~\cite{fang2021snunet} & \checkmark & & \checkmark & \checkmark & \\
    BIT~\cite{chen2021remote} & \checkmark & & & \checkmark & \\
    ChangeStar~\cite{zheng2021change} & \checkmark & & & & \\
    DSAMNet~\cite{shi2021deeply} & & & & \checkmark & \\
    P2V-CD~\cite{lin2022transition} & & & & \checkmark &\\
    ChangeFormer~\cite{bandara2022transformer} & \checkmark & & & & \\
    Changer~\cite{fang2022changer} & \checkmark & & & & \\
    TinyCD~\cite{codegoni2022tinycd} & \checkmark & & & &\\
    HANet~\cite{han2023hanet} & \checkmark & & & & \\
    LightCDNet~\cite{xing2023lightcdnet} & \checkmark & & & &\\
    CGNet~\cite{han2023change} & \checkmark & & & & \\
    BAN~\cite{li2024new} & \checkmark & & & &\\
    TTP~\cite{chen2023time} & \checkmark & & & & \\
    MTKD~\cite{liu2025jl1} & \checkmark & & & & \\
    
    \bottomrule[1pt]
  \end{tabular}}
  \vspace{-1em}
\end{table*}

\section{Supported methods}
\label{sec:supported_methods}

Open-CD contains high-quality implementations of popular change detection methods. A summary of supported frameworks and features compared with other codebases is provided in Table \ref{table_1}. A list is given as follows.

\subsection{Change Detection Methods}

\begin{itemize}
  \item \textbf{FC-EF}~\cite{daudt2018fully} follows U-Net architecture. It concatenates the bi-temporal images before passing them through the network, treating them as different color channels.

  \item \textbf{FC-Siam-Conc}~\cite{daudt2018fully} is a Siamese network. Compared to FC-EF, it separates the encoder layers into two streams with the same structure and shared weights (i.e., Siamese network) that process the bi-temporal images separately. In the skip connection, it concatenates the bi-temporal features and feeds them to the decoder.

  \item \textbf{FC-Siam-Diff}~\cite{daudt2018fully} is a variant of FC-Siam-Conc. Instead of concatenating both features from the encoding streams, it calculates the absolute value of their difference.

  \item \textbf{STANet}~\cite{chen2020spatial} is a metric-based change detection model. It consists of feature extractor, spatial-temporal attention module and metric module. The spatial-temporal attention module is a variant of self-attention in change detection task, which captures the global spatial–temporal relationships in the whole space-time to obtain more discriminative features. This attention module has two forms, basic spatial-temporal attention module (BAM) and pyramid spatial-temporal attention module (PAM). In addition, to reduce the effect of class imbalance, a class-sensitive loss, batch-balanced contrastive loss (BCL), is designed in STANet, which uses the number of samples in each class within a batch to determine the weight factor in the loss function.

  \item \textbf{DSIFN}~\cite{zhang2020deeply} builds on the FC-Siam-conc model. It uses channel and spatial attention to construct stronger basic modules (like CBAM) and also introduces deep supervision for fast convergence and performance improvement of the model.

  \item \textbf{SNUNet}~\cite{fang2021snunet} is a standard encoder-decoder architecture and uses the Siamese network as encoder. To maintain high-resolution features and fine-grained localization information, SNUNet uses the dense skip connection mechanism between the encoder and decoder (like UNet++). For fusing multi-granularity features, the ensemble channel attention module (ECAM) is designed to automatically select and focus on more effective information between different decoder groups. Structurally, ECAM is a natural expansion of plain channel attention module (CAM) in deep supervision and ensemble learning.

  \item \textbf{BIT}~\cite{chen2021remote} is a CNN-Transformer based change detction model. It abstracts pixel-level features into several visual tokens and models spatio-temporal context information in a compact token space. Compared with directly extracting dense spatio-temporal semantic correlations in pixel-level space, on the one hand, it reduces the interference of redundant information in image space through pixel-level feature aggregation, and uses Transformer encoders to construct spatio-temporal correlations, which significantly reduces computational complexity. The calculation efficiency is improved. On the other hand, the Transformer decoder is used to enhance the original pixel-level features with the learned context information, and the global spatio-temporal information is used to enhance the pixel-level representation, which improves the ability of the model to identify objects of interest and exclude irrelevant changes.
  
  \item \textbf{ChangeStar}~\cite{zheng2021change} and ChangeStar2~\cite{zheng2024single} are simple yet unified change detectors capable of addressing binary change detection, object change detection, and semantic change detection, which is composed of a Siamese dense feature extractor and ChangeMixin or ChangeMixin2. This design is inspired by reusing the modern semantic segmentation architecture because semantic segmentation and change detection are both dense prediction tasks. ChangeMixin and ChangeMixin2 enable any off-the-shelf deep semantic segmentation network to detect changes.

\footnotetext[9]{\url{https://github.com/likyoo/change_detection.pytorch}}
\footnotetext[10]{\url{https://github.com/ChenHongruixuan/ChangeDetectionRepository}}
\footnotetext[11]{\url{https://github.com/Bobholamovic/CDLab}}
\footnotetext[12]{\url{https://github.com/Bobholamovic/ChangeDetectionToolbox}}

  \item \textbf{ChangeFormer}~\cite{bandara2022transformer} is a variant of SegFormer~\cite{xie2021segformer} in change detection task. Different from SegFormer, it uses a Siamese customized mix transformer (MiT) as the encoder. The bi-temporal features in the Siamese network are concatenated and then fed into the decoder consisting of multi-layer perceptron (MLP) layers.

  \item \textbf{Changer}~\cite{fang2022changer} series models emphasize interactions between bi-temporal branches. There are two simple interaction strategies: aggregation-distribution (AD) and feature “exchange.” Specifically, the AD interaction is abstracted from some co-/cross-attention mechanisms. The ``exchange'' interaction is a completely parameter and computation-free operation, which is achieved by exchanging bi-temporal feature maps in the spatial or channel dimension, and the exchanged features are mixed as they pass through subsequent convolution or token mixer. In addition, the flow-based dual-alignment fusion (FDAF) module is proposed to overcome the problem of side-looking and misalignment in multi-temporal images. 

  \item \textbf{TinyCD}~\cite{codegoni2022tinycd} aims to develop a deep learning model that significantly reduces memory consumption and computational complexity. To achieve this, TinyCD extracts low-level features from two images using the early layers of a selected Siamese backbone. This approach minimizes the model's size and leverages the locality structure of features induced by the inductive bias inherent in the initial stages of a fully convolutional backbone.
  Subsequently, TinyCD introduces an attention module called MAMB. This module comprises an initial mixing stage that employs grouped convolution to integrate semantically similar features from the two images. Following this, a MLP fuses the mixed features into a single heatmap. This heatmap serves as a skip connection, reweighting the upsampled features during the upsampling phase of the Siamese U-Net. This process functions as a spatio-temporal attention mechanism.
  In the final stage, TinyCD utilizes a multi-layer perceptron as a pixel-level classifier to produce a probability score indicating change for each pixel.
  These architectural choices result in a model with a significantly reduced number of parameters and computational complexity while still achieving state-of-the-art performance. Consequently, TinyCD presents a valuable alternative to other leading models, particularly in scenarios where rapid training and deployment on low-end devices are critical.

  \item \textbf{HANet}~\cite{han2023hanet} is a discriminative Siamese network, hierarchical attention network, which can integrate multiscale features and refine detailed features. HANet has four progressive foreground-balanced sampling strategies based on not adding change information to help the model accurately learn the features of the changed pixels during the early training process and thereby improve detection performance. The main part of HANet is the HAN module, which is a lightweight and effective self-attention mechanism.

  \item \textbf{LightCDNet}~\cite{xing2023lightcdnet} is a lightweight and accurate change detection model designed for real-world application deployment. LightCDNet follows the standard ``encoder-decode'' structure, but unlike most Siamese network-based models, this model have designed a novel early fusion module to facilitate cross-temporal information fusion, called the deep supervised fusion module (DSFM), which effectively enhances the information preservation capability of the change detection model even with a small number of parameters. On the other hand, LightCDNet uses an improved ShuffleNetV2 as the encoder, coupled with a feature pyramid decoder to achieve high-precision change detection performance. LightCDNet provide versions with different parameter quantities for flexible deployment.

  \item \textbf{CGNet}~\cite{han2023change} is the change guiding network to tackle the insufficient expression problem of change features in the conventional U-Net structure adopted in previous methods. It contains a self-attention module named change guide module, which can effectively capture the long-distance dependency among pixels and effectively overcomes the problem of the insufficient receptive field of traditional convolutional neural networks.

  \item \textbf{BAN}~\cite{li2024new} is a universal foundation model-based change detection adaptation framework aiming to extract the knowledge of foundation models for change detection. It contains three parts, i.e. frozen foundation model (e.g., CLIP), bi-temporal adapter branch (Bi-TAB), and bridging modules between them. Specifically, BAN extracts general features through a frozen foundation model, which are then selected, aligned, and injected into Bi-TAB via the bridging modules. Bi-TAB is designed as a model-agnostic concept to extract task/domain-specific features, which can be either an existing arbitrary change detection model or some hand-crafted stacked blocks.

  \item \textbf{TTP}~\cite{chen2023time} enhances high-precision change detection in complex spatio-temporal remote sensing scenarios by integrating the general knowledge of foundational visual models into the change detection task. It leverages the potential of foundational models even with limited annotated change detection data. TTP overcomes domain shift issues during knowledge transfer and addresses the challenge of expressing both homogeneous and heterogeneous features in multi-temporal images. Specifically, TTP utilizes general segmentation knowledge based on the segment anything model (SAM) by introducing low-rank fine-tuning parameters into the SAM backbone, which mitigates spatial semantic domain shifts. Furthermore, TTP proposes a time-travel activation gate, enabling temporal features to permeate the pixel semantic space, thus enhancing the foundational model's ability to understand both homogeneous and heterogeneous features of bi-temporal images. Lastly, an efficient multi-level change prediction head is designed to decode dense and high-level semantic change features within the foundational model. This method paves the way for more accurate and efficient change detection in remote sensing image.

  \item \textbf{MTKD}~\cite{liu2025jl1} is an innovative approach to enhance change detection in remote sensing image. Existing algorithms often struggle to consistently deliver satisfactory results across the diverse scenarios of natural and human-induced changes. MTKD is designed to address this issue comprehensively. The training process of the MTKD framework is comprised of three stages. Initially, an original model is trained on the complete training set. Subsequently, the training data is partitioned based on the change area ratio (CAR), enabling the optimization of distinct teacher models for each specific range of changes. Finally, the student model, initialized with the parameters from the original model, is trained under the supervision of these specialized teacher models, integrating the diverse strengths taught by each teacher. This methodology significantly enhances the student's ability to manage a wide range of change scenarios. The MTKD framework empowers the student model to achieve superior detection accuracy without additional computational or time costs during inference, marking a significant advancement in change detection by ensuring robust performance across various network architectures.

\end{itemize}

\begin{figure*}[t]
  \centering
   \includegraphics[width=0.9\linewidth]{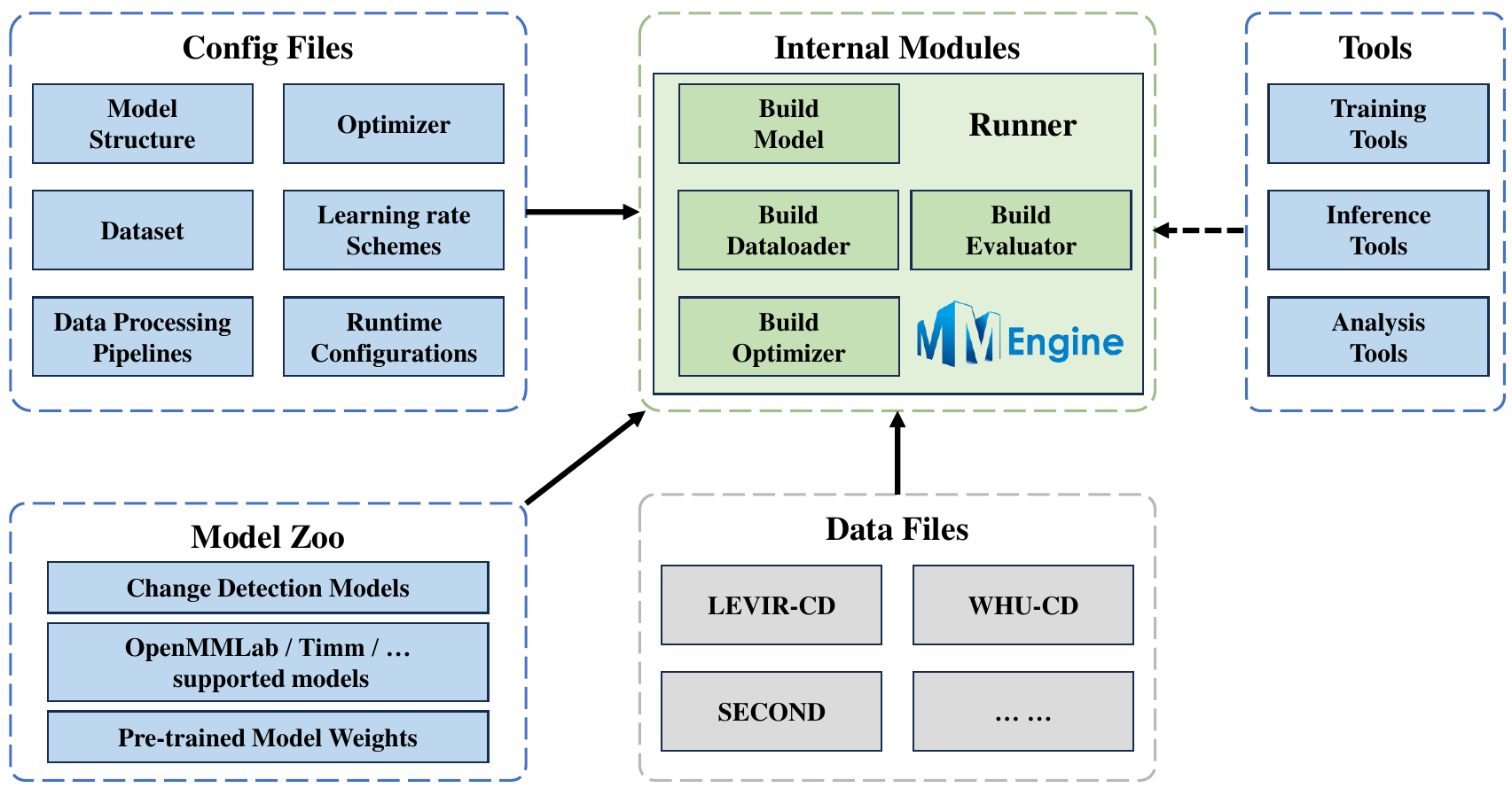}
   \caption{The overall architecture of Open-CD.}
   \label{fig:fig1}
\end{figure*}

\subsection{Customized Change Detection Models}

There is no doubt that the construction of change detection models benefits from other fundamental tasks in general vision, especially classification and segmentation. The former provides powerful feature extractors (backbones) and the latter provides extensive change mask generation schemes. Therefore, a core aspect of Open-CD's design is to allow users to arbitrarily call components from other vision models and to make it easier for change detection to benefit from advanced general modules. This is an important reason why we build Open-CD on top of OpenMMLab toolkits.

Another change detection codebase that allows for custom module combinations is change\_detection.pytorch (CDP), which is built on a "Siamese encoder + feature fusion + decoder" mode and supports some common backbone networks and segmentation heads. However, since CDP relies on developers to frequently code to support the latest models, it is difficult to maintain and update. Open-CD, on the other hand, is able to benefit directly from the constantly updated OpenMMLab Toolkits, and to modify and combine hundreds of models and modules. \textbf{{\color{green!50!black}In brief, Open-CD allows users to play free architecture games.}}

\section{Architecture}

\subsection{Overall}

\begin{table*}
  \caption{Supported datasets in Open-CD.}
  \label{table_2}
  \centering
  \scalebox{0.9}{
  \begin{tabular}{@{}lcccccc@{}}
    \toprule[1pt]
    Dataset & Task & Images pairs & Image size & Change instances & Change pixels & Download \\
    \midrule[1pt]
    LEVIR-CD~\cite{chen2020spatial} & binary & 637 & 1024 $\times$ 1024 & 31K & 30M & \href{https://chenhao.in/LEVIR/}{URL}\\
    WHU-CD~\cite{ji2018fully} & binary & 1 & 32,207 $\times$ 15,354 & 2K & 21M & \href{https://study.rsgis.whu.edu.cn/pages/download/building_dataset.html}{URL} \\
    S2Looking~\cite{shen2021s2looking} & binary & 5,000 & 1024 $\times$ 1024 & 66K & 69M & \href{https://github.com/S2Looking/Dataset}{URL} \\
    SVCD~\cite{lebedev2018change} & binary & 16,000 & 256 $\times$ 256 & - & - & \href{https://drive.google.com/file/d/1GX656JqqOyBi_Ef0w65kDGVto-nHrNs9/edit}{URL} \\
    DSIFN~\cite{zhang2020deeply} & binary & 394 & 512 $\times$ 512 & - & - & \href{https://github.com/GeoZcx/A-deeply-supervised-image-fusion-network-for-change-detection-in-remote-sensing-images/tree/master/dataset}{URL} \\
    CLCD~\cite{liu2022cnn} & binary & 560 & 512 $\times$ 512 & - & - & \href{https://github.com/liumency/CropLand-CD}{URL} \\
    RSIPAC~\cite{RSIPAC} & binary & 3,194 & 512 $\times$ 512 & - & - & \href{https://engine.piesat.cn/ai/autolearning/index.html\#/dataset/detail?key=8f6c7645-e60f-42ce-9af3-2c66e95cfa27}{URL} \\
    JL1-CD~\cite{liu2025jl1} & binary & 5,000 & 512 $\times$ 512 & 13K & 127M & \href{https://github.com/circleLZY/MTKD-CD}{URL} \\
    SECOND~\cite{yang2020semantic} & semantic & 4,662 & 512 $\times$ 512 & - & - & \href{http://www.captain-whu.com/PROJECT/}{URL} \\
    Landsat~\cite{yuan2022transformer} & semantic & 8,468 & 416 $\times$ 416 & - & - & \href{https://figshare.com/articles/figure/Landsat-SCD_dataset_zip/19946135/1}{URL} \\
    BANDON~\cite{pang2023detecting} & semantic & 2,283 & 2048 $\times$ 2048 & 123K & 283M & \href{https://github.com/fitzpchao/BANDON}{URL} \\
    \bottomrule[1pt]
  \end{tabular}}
  \vspace{-1em}
\end{table*}







As shown in Figure \ref{fig:fig1}, Open-CD consists of five parts.

\textbf{Config Files.} In Open-CD, the model, optimizer, dataset, data pipeline, etc. are specified in config files. In general, if the user does not need to redesign the model internally, simply modifying the config file is enough.

\textbf{Model Zoo.} Besides change detection models listed in Table \ref{table_1}, Open-CD supports almost all models in OpenMMLab related algorithm codebases, which can be called directly in the config file by simply installing them as dependencies. In addition, there are thousands of pre-trained weights available, including plain ImageNet, foundation models, etc.

\textbf{Internal Modules.} Following the design concept of the OpenMMLab toolkits, the training process of Open-CD is conducted in the MMEngine Runner. We will describe it in Section \ref{sec:training_pipeline}.

\textbf{Data Files.} Open-CD supports the current mainstream binary and semantic change detection datasets, as listed in Table \ref{table_2}. And since we define two extensible base classes, \texttt{\_BaseCDDataset} and \texttt{BaseSCDDataset}, it is easy for users to integrate custom datasets into Open-CD.

\textbf{Tools.} Open-CD also contains several functional scripts e.g. training tools, inference tools, data analysis tools, result analysis tools, etc.

\subsection{Training Pipeline}
\label{sec:training_pipeline}

\begin{figure}[t]
  \centering
   \includegraphics[width=1.0\linewidth]{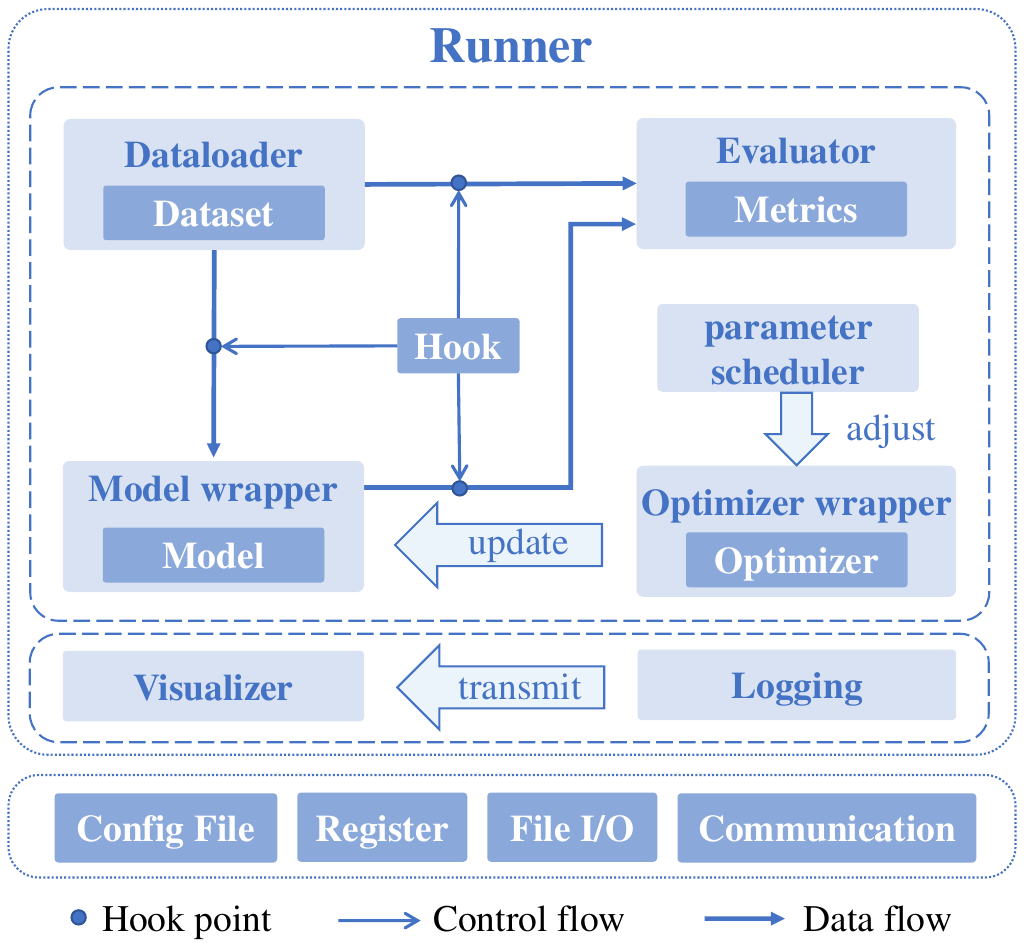}
   \caption{Training Pipeline of Open-CD. \protect\footnotemark}
   \label{fig:fig2}
\end{figure}

Similar to other algorithm codebases in OpenMMLab toolkits, Open-CD follows the unified interface of abstract components defined in MMEngine. The training pipeline of Open-CD is shown in Figure \ref{fig:fig2}. First, the Dataloader obtains data from the dataset and transmits it to the model. Outside the model, there is a Wrapper for distributed training, etc. Then, the parameter scheduler is used to adjust Optimizer related parameters e.g. learning rate during training, and the Optimizer is used to update the model. The Optimizer wrapper is capable of gradient accumulation, gradient clipping, mixed precision training, etc. Finally, in the evaluation phase, the data and model outputs are fed to the Evaluator and the corresponding values are returned based on the evaluation metrics.

To make the training pipeline extensible, some hook points are inserted into the pipeline. With these hooks, the data flow in the Runner can be operated and observed to achieve the user-customized requirements. During the whole training pipeline, the Logging component is constantly working, transmitting the required information to the Visualizer.

\footnotetext[13]{\url{https://github.com/open-mmlab/OpenMMLabCourse}}
\section{Benchmarks}

\subsection{Experimental Setting}

\noindent\textbf{Dataset.}
We adopt LEVIR-CD as the primary benchmark for all experiments because it has high quality and is more widely used. We use the \texttt{Train} split for training and report the performance on the \texttt{Test} split.

\noindent\textbf{Implementation details.}
If not otherwise specified, we adopt the following settings:

\begin{enumerate}[label=(\arabic*)]
    \item Images are randomly cropped to 256 $\times$ 256.
    \item We use a single RTX 3090 GPU for training and inference, with a training batch size of 8.
    \item The training schedule is ``40k'', meaning 40k iteration.
    \item Data augmentation including: \texttt{RandomRotate}, \texttt{RandomFlip} and \texttt{PhotoMetricDistortion}.
\end{enumerate}

\noindent\textbf{Evaluation metrics.}
We adopt $F_1^c$, $IoU^c$, $Precision^c$ and $Recall^c$ as evaluation metrics. Note that \textbf{{\color{green!50!black}all metrics are based on the category ``change''}} to avoid the class imbalance problem.

\subsection{Benchmarking Results}

\begin{table*}
  \caption{Results of different change detection methods on LEVIR-CD \texttt{Test}. \dag \ denotes the size of random crop is $512 \times 512$, * denotes the batch size is set to 16, \ddag \ denotes the official code is based on Open-CD, and ``300e'' denotes 300 epochs. For BAN and TTP, only the learnable parameters are counted.}
  \label{table_levircd}
  \centering
  \scalebox{0.9}{
  \begin{tabular}{@{}l|c|c|c|c|c|c|c|c|c@{}}
    \toprule[1pt]
    Method & Backbone & Lr Schd & Param (M) & GFLOPs & Inf (fps) & $Precision^c$ & $Recall^c$ & $F_1^c$ & $IoU^c$\\
    \hline
    FC-EF~\cite{daudt2018fully} & - & 40k & 1.353 & 3.244 & 66.37$\pm$0.0333 & 80.54 & 81.17 & 80.85 & 67.86 \\
    FC-Siam-Diff~\cite{daudt2018fully} & - & 40k & 4.385 & 1.352 & 46.29$\pm$0.0015 & 89.64 & 80.25 & 84.68 & 73.44 \\
    FC-Siam-Conc~\cite{daudt2018fully} & - & 40k & 4.989 & 1.548 & 45.67$\pm$0.0016 & 86.59 & 84.53 & 85.55 & 74.75 \\
    STANet-PAM~\cite{chen2020spatial} & ResNet-18 & 40k & 13.356 & 48.083 & 1.53$\pm$0.0001 & 84.28 & 90.19 & 87.13 & 77.20 \\
    DSIFN~\cite{zhang2020deeply}& VGG-16 & 40k & 35.995 & 78.982 & 5.19$\pm$0.0001 & 89.78 & 92.32 & 91.03 & 83.54 \\
    SNUNet-c16~\cite{fang2021snunet} & - & 40k & 3.012 & 11.730 & 5.96$\pm$0.0001 & 93.02 & 89.75 & 91.36 & 84.09 \\
    BIT~\cite{chen2021remote} & ResNet-18 & 40k & 2.990 & 8.749 & 28.93$\pm$0.0006 & 93.17 & 88.38 & 90.71 & 83.00 \\
    ChangeStar\dag*~\cite{zheng2021change} & ResNet-18 & 40k & 16.965 & 19.213 & 27.06$\pm$0.0007 & 93.75 & 88.90	& 91.26 & 83.92 \\
    ChangeFormer-b0~\cite{bandara2022transformer} & MiT-b0 & 40k & 3.847 & 2.455 & 25.26$\pm$0.0005 & 93.07 & 88.20 & 90.57 & 82.76 \\
    ChangeFormer-b1~\cite{bandara2022transformer} & MiT-b1 & 40k & 13.941 & 5.825 & 18.71$\pm$0.0004 & 93.09 & 89.26 & 91.14 & 83.71 \\
    Changer\dag\ddag~\cite{fang2022changer} & ResNet-18 & 40k & 11.391 & 5.955 & 49.31$\pm$0.2815 & 92.86 & 90.78 & 91.81 & 84.86 \\
    TinyCD~\cite{codegoni2022tinycd} & - & 40k & 0.285 & 1.448 & 26.28$\pm$0.0004 & 91.87 & 89.89 & 90.87 & 83.26 \\
    HANet~\cite{han2023hanet} & - & 40k & 3.028 & 20.822 & 4.80$\pm$0.0001 & 91.72 & 89.09 & 90.39 & 82.46 \\
    LightCDNet-base\ddag~\cite{xing2023lightcdnet} & - & 40k & 1.313 & 3.164 & 17.60$\pm$0.0041 & 90.95 & 90.30 & 90.62 & 82.86 \\
    LightCDNet-large\ddag~\cite{xing2023lightcdnet} & - & 40k & 2.816 & 5.691 & 10.29$\pm$0.0002 & 92.68 & 89.67 & 91.15 & 83.74 \\
    CGNet~\cite{han2023change} & VGG-16 & 40k & 38.989 & 87.550 & 1.93$\pm$0.0001 & 93.60 & 90.64 & 92.10 & 85.36 \\
    BAN (ViT-L)\dag\ddag~\cite{li2024new} & MiT-b0 & 40k & 4.474 & 307.374 & 1.87$\pm$0.0001 & 93.47 & 90.30 & 91.86 & 84.94 \\
    TTP\dag\ddag~\cite{chen2023time} & ViT-L & 300e & 6.210 & 929.840 & 0.67$\pm$0.0001 & 93.00 & 91.70 & 92.10 & 85.60 \\
    \bottomrule[1pt]
  \end{tabular}}
\end{table*}


\textbf{Main results.} We benchmark different methods on LEVIR-CD \texttt{Test}, including FC-EF~\cite{daudt2018fully}, FC-Siam-Diff~\cite{daudt2018fully}, FC-Siam-Conc~\cite{daudt2018fully}, STANet-BAM/PAM~\cite{chen2020spatial}, DSIFN~\cite{zhang2020deeply}, SNUNet-c16~\cite{fang2021snunet}, BIT~\cite{chen2021remote}, ChangeStar~\cite{zheng2021change}, ChangeFormer-b0/b1~\cite{bandara2022transformer}, Changer~\cite{fang2022changer}, TinyCD~\cite{codegoni2022tinycd}, HANet~\cite{han2023hanet}, LightCDNet-base/large~\cite{xing2023lightcdnet}, CGNet~\cite{han2023change}, BAN (ViT-L)~\cite{li2024new} and TTP~\cite{chen2023time}. We report the parameter, GFLOPs, inference speed, $Precision^c$, $Recall^c$, $F_1^c$ and $IoU^c$ of these methods in Table \ref{table_levircd}. The inference time is tested on a single RTX 3090 GPU. The GFLOPs is calculated at a size of $256 \times 256$. For re-production and comparison in subsequent studies, we will also release all \textbf{{\color{green!50!black}weights}} and \textbf{{\color{green!50!black}visualization results}}.

\begin{figure}[t]
  \centering
   \includegraphics[width=1.0\linewidth]{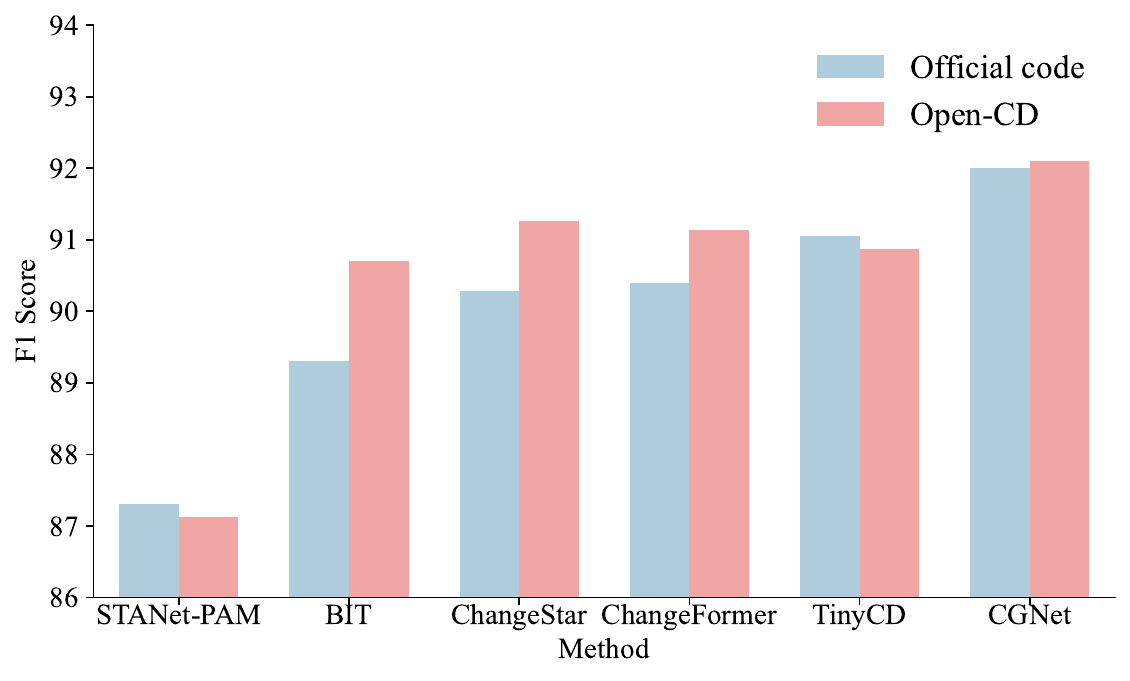}
   \caption{Comparison with official code.}
   \label{fig:fig3}
\end{figure}

\textbf{Comparison with official code.} We compare the official implementations of some methods with those in Open-CD, as shown in Figure \ref{fig:fig3}. With the exception of STANet and TinyCD, all other implementations in Open-CD gain some degree of performance improvement compared to the original implementations. Especially for BIT, ChangeStar and ChangeFormer, the improvements are significant. Notably, the original ChangeFormer customizes the configuration of MiT (e.g. depth, width of the network, etc.). In Open-CD, we restore the settings of b1$\sim$b5 in SegFormer to make it scalable, and obtain higher $F_1^c$ score with fewer parameters.

\section{Application}
  
\subsection{Algorithm implementation, validation and deployment}

As mentioned above, Open-CD can be used for the implementation of custom change detection algorithms, validation of existing algorithms, and model deployment (relying on the OpenMMLab ecosystem, i.e., MMDeploy\footnote{\url{https://github.com/open-mmlab/mmdeploy}}. Although the current Open-CD only contains supervised learning pipelines, it can also be applied as a construction tool for models, datasets, etc., promoting the exploration of semi-supervised and unsupervised algorithms.

\subsection{Downstream validation for foundation model}

Recently, researchers tend to explore the general capabilities of foundation models. Within the remote sensing community, foundation models are also actively explored. Some works attempt to train a robust backbone that is capable of extracting general features for a series of downstream perceptual tasks e.g., object detection, semantic segmentation, change detection, etc. For this, Open-CD is an out-of-the-box downstream validation tool for change detection. Typically, MTP~\cite{MTP}, a remote sensing foundation model via multi-task pre-training, yields an image encoder for downstream task fine-tuning and uses Open-CD for validation of change detection task.

\subsection{Downstream validation for generative model}

Although huge amount of remote sensing images are available through Google Earth etc., data annotation remains a challenge, which is time-consuming and costly. One trend is to synthesize simulated image-label pairs through conditional GAN, Diffusion models, etc.~\cite{zheng2023scalable, tang2024changeanywhere, zheng2024changen2} and use these data to (pre-)train downstream perception models. In this case, Open-CD can be applied to validate the quality of the simulated change detection data. A typical study is ChangeAnywhere~\cite{tang2024changeanywhere}, which proposes a diffusion model to generate a simulated change detection dataset ChangeAnywhere-100K and uses Open-CD for subsequent validation.

\subsection{Competition and challenge}

As an advanced algorithm codebase, Open-CD can be easily used for change detection competitions and challenges. In general, users only need to customize their \texttt{Dataset} Class to enjoy all the advanced algorithms, data analysis tools, inference tools, etc. in Open-CD. Here, some solutions of winning competitions with Open-CD are listed:

\begin{itemize}
  \item \textbf{2024 ``Jilin-1'' Cup.\footnote{\url{https://www.jl1mall.com/contest}}}
  \textbf{{\color{green!50!black}2nd place.}}
  \footnote{\url{https://github.com/circleLZY/MTKD-CD}}
  
  \item \textbf{2024 ISPRS TC 1 Contest.\footnote{\url{https://www.gaofen-challenge.com/challenge}}}
  \textbf{{\color{green!50!black}Winner.}}
  \footnote{\url{https://github.com/DanyangLihhh/2024-ISPRS-TC-I-Contest}}

  \item \textbf{2023 ``Jilin-1'' Cup.}
  \textbf{{\color{green!50!black}5th place.}}
  \footnote{\url{https://github.com/DanyangLihhh/Cultivated-land-change-detection}}
\end{itemize}

\section{Acknowledgements}

The authors would like to express their deep gratitude to the OpenMMLab team for their invaluable contributions to vision community. The authors would also like to extend their appreciation to the MMDetection team for their excellent technical report, which provided a valuable reference for the writing of this report.
{
    \small
    \bibliographystyle{ieeenat_fullname}
    \bibliography{main}
}


\end{document}